\documentclass[11pt,a4paper]{article}
\usepackage[hyperref]{emnlp-ijcnlp-2019}
\usepackage{times}
\usepackage{latexsym}
\usepackage{textgreek}
\usepackage{mathtools}
\usepackage{bbm}
\usepackage{float}
\usepackage{enumerate}
\usepackage{amsmath,amsfonts,amsthm,bm} 
\usepackage{url}
\renewcommand*{\thefootnote}{\fnsymbol{footnote}}
\aclfinalcopy 

\begin{document}
\title{Technical report on Conversational Question Answering}

\author{Ying Ju,\; Fubang Zhao,\; Shijie 
    Chen\footnotemark{},\; Bowen Zheng\footnotemark[\value{footnote}],\\
         Xuefeng Yang,\; Yunfeng Liu \\\\
        ZhuiYi Technology, Shenzhen, China \\
        \tt \{bettyju, nicolaszhao, ryan, glenliu\}@wezhuiyi.com \\
        \tt chrischen369@outlook.com \\
        \tt zaberchann@gmail.com 
        }
\date{}
\maketitle{}

\begin{abstract}
Conversational Question Answering is a challenging task since it requires understanding of conversational history. In this project, we propose a new system RoBERTa + AT + KD, which involves rationale tagging multi-task, adversarial training, knowledge distillation and a linguistic post-process strategy. Our single model achieves \textbf{90.4} (F1) on the CoQA test set without data augmentation, outperforming the current state-of-the-art single model by \textbf{2.6\%} F1.

\footnotetext[1]{This work was done while the author was doing an internship at ZhuiYi Technology}
\renewcommand*{\thefootnote}{\arabic{footnote}}
\end{abstract}

\section{Introduction}
Conversational question answering (CQA) is a QA task involving comprehension of both passage and historical QAs.  It is proposed to mimic the way human seek information in conversations. Recently, several CQA datasets were proposed, such as CoQA\cite{2018arXiv180807042R}, QuAC \cite{2018arXiv180807036C} and QBLink \cite{elgohary-etal-2018-dataset}.

In this paper, we propose RoBERTa + AT + KD, a system featuring Adversarial Training (AT) and Knowledge Distillation (KD) for CQA tasks. Empirical results on the CoQA dataset show the effectiveness of our system, which outperforms the previous best model by a \textbf{2.6\%} absolute improvement in F1 score. 


The contributions of our paper are as follows:
\begin{itemize}
\item We propose a general solution to fine-tuning pre-trained models for CQA tasks by (1) rationale tagging multi-task utilizing the valuable information in the answer's rationale; (2) adversarial training \cite{2016arXiv160507725M} increasing model's robustness to perturbations; (3) knowledge distillation \cite{2018arXiv180504770F} making use of additional training signals from well-trained models.
\item We also analyze the limitation of extractive models including our system. To figure out the headroom of extractive models for improvement, we compute the proportion of QAs with free-form answers and estimate the upper bound of extractive models in F1 score.
\item Our system achieves the new state-of-the-art result on CoQA dataset without data augmentation.
\end{itemize}

\section{Related Work}
\textbf{Machine Reading Comprehension} Machine Reading Comprehension(MRC) has become an important topic in natural language processing. Existing datasets can be classified into single-turn or multi-turn according to whether each question depends on the conversation history. Many MRC models have been proposed to tackle single-turn QA, such as BiDAF\cite{2016arXiv161101603S}, DrQA\cite{2017arXiv170400051C}, R-Net\cite{rnet} and QANet\cite{2018arXiv180409541Y} in SQuAD dataset. For multi-turn QA, existing models include FlowQA\cite{2018arXiv181006683H} and SDNet\cite{2018arXiv181203593Z}. FlowQA proposed an alternating parallel processing structure and incorporated intermediate representations generated during the process of answering previous questions. SDNet introduced an innovated contextualized attention-based deep neural network to fuse context into traditional MRC models.

\textbf{Pretrained Model} Pretrained language models, such as GPT\cite{gpt}, BERT\cite{2018arXiv181004805D}, XLNET\cite{2019arXiv190608237Y} and RoBERTa\cite{2019arXiv190711692L}, have brought significant performance gains to many NLP tasks, including machine reading comprehension. BERT is pretrained on ``masked language model'' and ``next sentence prediction'' tasks. RoBERTa can match or exceed the performance of all post-BERT methods by some modifications. Many recent models targeting CoQA are based on BERT, such as Google SQuAD 2.0 + MMFT, ConvBERT and BERT + MMFT + ADA\footnote[1]{https://stanfordnlp.github.io/coqa/}, but there is no model based on RoBERTa until now.


\section{Methods}
In this section, we first introduce a baseline model based on RoBERTa(a Robustly optimized BERT pretrainig Approach) for the CoQA dataset, and then adopt some methods to improve the model performance.    
\subsection{Baseline: RoBERTa adaption to CoQA}

Different from other QA datasets, questions in CoQA are conversational. Since every question after the first one depends on the conversation history, each question is appended to its history QAs, similar to \cite{2018arXiv180807042R} . If the question in the $k^{th}$ turn is $Q_k$, the reformulated question is defined as:  
\begin{equation}
Q_k^*= \{Q_1,A_1,...,Q_{k-1},A_{k-1},Q_k\}
\end{equation}
Symbol $[Q]$ and symbol $[A]$ are added before each question and each answer respectively. To make sure the length of $Q_k^*$ is within the limit on the number of question tokens, we truncate it from the end. 
In our experiment, the pre-trained model RoBERTa is employed as our baseline model, which takes a concatenation of two segments as input. Specifically, Given a context ${C}$, 
the input $x$ for RoBERTa is [CLS] $Q_k^*$ [SEP] ${C}$ [SEP]. 

The answers of CoQA dataset can be free-form text, Yes, No or Unknown. Except the Unknown answers, each answer has its rationale, an extractive span in the passage. Considering that, we adopt an extractive approach with Yes/No/Unk classification to build the output layer on top of RoBERTa. First, the text span with the highest f1-score in rationale is selected as the gold answer for training. Then we use a FC layer to get the start logits $l^s$ and end logits $l^e$. For Yes/No/Unk classification, a FC layer is applied to the RoBERTa pooled ouptut $h^p$ to obtain three logits $l^y$, $l^n$ and $l^u$. The objective function for our baseline model is defined as:
\begin{align}
    p^{s} &= softmax([l^s,\; l^y,\; l^n,\; l^u]) \\
    p^{e} &= softmax([l^e,\; l^y,\; l^n,\; l^u])  \\
    \mathcal{L}_{Base} &= -\frac{1}{2N}\sum_{i=1}^N (\log{p_{y_{i}^s}^{s}}+\log{p_{y_{i}^e}^{e}})
\end{align}
where $y_i^s$ and $y_i^e$ are starting position and ending position in example $i$ respectively, and the total number of examples is $N$.


\subsection{Rationale Tagging Multi Task}
As mentioned above, every answer except Unknown has its rationale. To utilize this valuable information, we add a new task Rationale Tagging in which the model predicts whether each token of the paragraph should be included in the rationale. In other words, tokens in the rationale will be labeled 1 and others will be labeled 0. For Unk questions, they should all be 0.

Therefore, besides predicting the boundary and classification, we also add an extra FC layer to compute the probability for rationael tagging:
\begin{equation}
       p_t^r = {sigmoid}({w_{2}}{Relu}({W}_{1}{h}_t))
\end{equation}%
where ${h}_t\in\mathbb{R}^d$ is the RoBERTa's output vector for $t^{th}$ token, ${W_1}\in\mathbb{R}^{d\times d}$, ${w_2}\in\mathbb{R}^d$. The rationale loss is defined as the averaged cross entropy.
\begin{align}
    \mathcal{L_\text{RTi}}&=-\frac{1}{T}\sum_{t=1}^{T}({y_{it}^r}{\log}{p_{it}^r}+(1-y_{it}){\log}({1-p_{it}^r}))\\
    \mathcal{L_{\text{RT}}} &= \frac{1}{N}\sum_{i=1}^{N}\mathcal{L_\text{RTi}}
\end{align}

Here ${T}$ is the number of tokens, ${y_t^r}$ is the rationale label for the ${t}^{th}$ token and $L_{RTi}$ is rationale loss for the $i^{th}$ example. The model is trained by jointly minimizing the two objective functions:
\begin{equation}
    \mathcal{L} = \mathcal{L_{\text{Base}}} + \beta_{1}\mathcal{{L_{\text{RT}}}}
\end{equation}
where $\mathbf{\beta_1}$ is a hyper-parameter for weights of rationale loss.

%
%

In addition, rationales can be used to assist the classification of Yes/No/Unk questions. The process works as follows: We first compute the rationale sequence output ${h_t^r}$ by multiplying ${p_t^r}$ and ${h_t}$. Then a attention layer for ${h_t^r}$ is used to obtain the rationale representation $h^{p*}$.
\begin{align}
    {h_t^r} &= p_t^r \times h_t \\
    {a_t} &= {softmax}_{t=1}^T(w_{a2}Relu(W_{a1}h_t^r)) \\
    {h^{p*}} &= \sum_{t=1}^{T}{a_t}\times{h_t} 
\end{align}
where $W_{a1}\in\mathbb{R}^{d\times d}$, $w_{a2}\in\mathbb{R}^d$ are learned parameters. Finally, when producing the $l^y$, $l^n$ and $l^u$ for Yes/No/Unk, we replace $h^p$ with $\big[{h}^{p*}\;{h}^p\big]$, which is the concatenation of RoBERTa's pooled output $h^p$ and the rationale representation $h^{p*}$.

\subsection{Adversarial and Virtual Adversarial Training}
Adversarial training \cite{2014arXiv1412.6572G} is a regularization method for neural networks to make the model more robust to adversarial noise and thereby improves its generalization ability.

\textbf{Adversarial Training (AT)} In this work, adversarial examples are created by making small perturbations to the embedding layer. Assuming that ${v}_w$ is the embedding vector of word $w$ and ${\hat{\theta}}$ represents the current parameters of the model, the adversarial embedding vector ${v}^*_w$ \cite{2016arXiv160507725M} is:
\begin{align}
{g}_w &= -\bigtriangledown_{{v}_w} {\mathcal{L}}(y_i|{v}_w; {\hat{\theta}} ) \\
{v}^*_w &= {v}_w+ \epsilon{g}_w / \|{g}_w\|_2
\end{align}
where $y_i$ is the gold label, $\epsilon$ is a hyperparameter scaling the magnitude of perturbation. Then, we compute adversarial loss as:
\begin{equation}
    \mathcal{L}_{\text{AT}}({\theta}) = -\frac{1}{N}\sum_{i=1}^{N}{CrossEntropy}(\cdot | {V}^*; {\theta})
\end{equation}
where ${V}^* = [v^*_{w_1}, ..., v^*_{w_n}]$ is the adversarial embedding matrix.

\textbf{Virtual Adversarial Training (VAT)} 
Virtual adversarial training is similar to AT, but it uses unsupervised adversarial perturbations. To obtain the virtual adversarial perturbation, we first add a gaussian noise to the embedding vector of word $w$:
\begin{equation}
{v}'_w = {v}_w + \xi {d}_w
\end{equation}
where $\xi$ is a hyperparameter and ${d}_w \sim \mathcal{N}(0, I)\in \mathbb{R}^d$. Then the gradient with respect to the KL divergence between $p(V)$ and $p(V')$ is estimated as:
\begin{equation}
    {g}_w = \bigtriangledown_{{v}'}D_{\text{KL}}(p(\cdot | {v}_w; {\hat{\theta}}) \| p(\cdot | {v}'_w; {\hat{\theta}}))
\end{equation}
Next, similar to adversarial training, the adversarial perturbation is added to the word embedding vector:
\begin{equation}
    {v}^*_w = {v}_w + \epsilon{g}_w / \|{g}_w\|_2
\end{equation}
Lastly, the virtual adversarial loss is computed as:
\begin{equation}
\mathcal{L}_{\text{VAT}}({\theta}) = \frac{1}{N}\sum^N_{i=1}D_{\text{KL}}(p(\cdot | {V}; {\theta}) \| p(\cdot | {V}^*; {\theta}))
\end{equation}
where ${V}^*$ is the adversarial embedding matrix.

\textbf{Loss Function} 
In this work, the total loss is simply summed up all the loss together as:
\begin{equation}
    \mathcal{L} = \mathcal{L}_{\text{BD}} + \beta_{1}\mathcal{L}_{\text{RA}} +\beta_{2}\mathcal{L}_{\text{AT}} + \beta_{3}\mathcal{L}_{\text{VAT}}
\label{total_loss}
\end{equation}

\subsection{Knowledge Distillation}
Knowledge Distillation(KD) \cite{2018arXiv180504770F} transfer "knowledge" from one machine learning model (teacher) to another (student) by using teacher's output as student's training objective. Even when the teacher and student share the same architecture and size, the student still outperforms its teacher. KD can be used for both single-task and multi-task models \cite{2019arXiv190704829C}.

\textbf{Teacher Model}
In this work, the teacher model is trained using methods mentioned above, whose objective function $\mathbf{\mathcal{L_\mathcal{T}}}$ is defined as Equation \ref{total_loss}.

\textbf{Student Model}
Knowledge distillation uses teacher's output probability ${{f}\big({x^i,\theta^\tau}\big)}$ as an extra supervised label to train student models. In our work, we employ several teacher models with different random seeds to compute the teacher label ${p_i^{kd}}$:
\begin{equation}
   {p_i^{kd}} = \frac{1}{\mathcal{T}}\sum^\mathcal{T}_{\tau=1}{f\big({x^i,\theta^\tau}\big)}
\end{equation}
where $\theta_\tau$ is the parameters of the $\tau^{th}$ teacher model and $\mathcal{T}$ is the total number of teacher models. KD loss $\mathcal{L}_{KD}$ is defined as the cross-entropy between ${p_i^{kd}}$ and ${{f}\big({x^i,\theta^s}\big)}$:
\begin{equation}
        \mathcal{L_{\text{KD}}}(\theta^s) = -\frac{1}{NT}\sum_{i=1}^N\sum_{t=1}^T{{p_{it}^{kd}}}{\log}f\big(x_{it},\theta^s\big) 
\end{equation}
Where $\mathbf{\theta^s}$ is the student model parameters. The total loss for student model is defined as:
\begin{equation}
        \mathcal{L_\text{S}} = \mathcal{L}_{\text{BD}} + \beta_{1}\mathcal{L}_{\text{RA}} +\beta_{2}\mathcal{L}_{\text{AT}} + \beta_{3}\mathcal{L}_{\text{VAT}} + \beta_{4}\mathcal{L}_{\text{KD}}
\end{equation}

\subsection{Post-Processing}
Since our model is extractive, it cannot solve multiple-choice questions even when it can extract the correct span. There are also multiple-choice questions whose options are the same as the word that our model finds but in a different word form. For instance, the options are 'walk' and 'ride' while the span that our model extracts is 'walked'.


A post-processing procedure based on word similarity is applied to alleviate this problem. 
First, legal options are extracted from questions using linguistic rules. Second, word embeddings of options and answer tokens are prepared respectively. Third, we compute the cosine similarity between each option and answer token. At last, the option with the highest similarity is chosen as the final answer. 
\begin{equation}
ans = \mathop{\arg\max}_{o} \ \left\{sim(o,a)|o\in\mathbb{O}, a\in\mathbb{A} \right\}
\end{equation}
where $\mathbb{O}$ and $\mathbb{A}$ are the set of word embeddings of option and answer tokens and $sim(o,a)$ represents their cosine similarity.

\subsection{Ensemble}
The ensemble output is generated according to the ensemble logits $L_e=\{L_e^s, L_e^e, L_e^y, L_e^n, L_e^u\}$, which is the average of output logits $l_{j}=\{l_e^s,l_e^e, l_e^y, l_e^n, l_e^u\}$ from models selected for ensemble.
\begin{equation}
    L_{e}=\frac{1}{M}\sum_{j=1}^{M}l_{j}
\end{equation}
where $M$ is ensemble size, the number of models used for ensemble.

The performance of our ensemble strategy relies heavily on the proper selection of models which is challenging. Given constraints on computational resources, the ensemble size is also limited. Genetic algorithm(GA), a kind of stochastic search algorithm that does not require gradient information, is used to search for a combination of models that maximizes performance while obeying the constraints on ensemble size. GAs are inherently parallel and tend to approximate the global optimal solution. Therefore they are widely used in combinatorial optimization and operation research problems\cite{deb1998genetic, muhlenbein1988evolution}. 


In our experiments, the genetic algorithm is able to converge within 200 generations. Our best ensemble solution contains only 9 models and reaches a F1 of 91.5 without post-processing while simply averaging all 68 candidate models results in a lower F1 of 91.2.
\begin{figure}[H]
   \centering
   \includegraphics[width=0.45\textwidth]{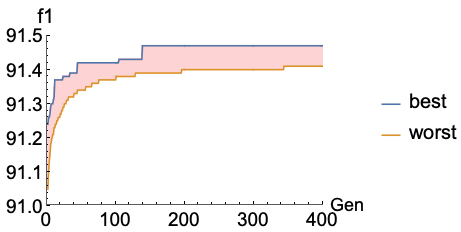}
 \caption{F1 score of best and worst individual of each generation.}\label{fig:ga_ensemble}
\end{figure}

\begin{table*}[t!]
\centering
\begin{tabular}{l|ccc}
\hline \bf Model & \bf In-domain & \bf Out-of-domain & \bf Overall \\ \hline
Bert-Large Baseline               & 82.6 & 78.4 & 81.4 \\
BERT with History Augmented Query & 82.7 & 78.6 & 81.5 \\
Bert + Answer Verification        & 83.8 & 81.9 & 82.8 \\
BERT + MMFT + ADA                 & 86.4 & 81.9 & 85.0 \\
ConvBERT                          & 87.7 & 85.4 & 86.8 \\
Google SQuAD 2.0 + MMFT           & 88.5 & 86.0 & 87.8 \\
Our model                         & 90.9 & \bf 89.2 & 90.4 \\
\hline
Google SQuAD 2.0 + MMFT(Ensemble) & 89.9 & 88.0 & 89.4 \\
Our model(Ensemble)               & \bf 91.4 & \bf 89.2 & \bf 90.7 \\
\hline
human                             & 89.4 & 87.4 & 88.8  \\
\hline
\end{tabular}
\caption{\label{result} CoQA test set results, which are scored by the CoQA evaluation server. All the results are obtained from https://stanfordnlp.github.io/coqa/.}
\end{table*}

\section{Experiments}
\subsection{Evaluation Metrics}
Similar to SQuAD, CoQA uses macro-average F1 score of word overlap as its evaluation metric. In evaluation, to compute the performance of an answer $a_p$, CoQA provides $n$ human answers $a_r^i$ as reference answers ($i\in\{1,2,...,n\}-, n=4$) and the final F1 score is defined as:
\begin{equation}
    {F1}_{{final}} = \frac{1}{n} \sum_{j=1}^n \max_{i=1}^n(\mathbbm{1}(i\ne j){F1}(a_p, a_r^i))
\end{equation}


\subsection{Implementation Details}
The implementation of our model is based on the PyTorch implementation of RoBERTa \footnote[1]{https://github.com/pytorch/fairseq/tree/master\\/examples/roberta}. We use AdamW \cite{2017arXiv171105101L} as our optimizer with a learning rate of 5e-5 and a batch size of 48. The number of epochs is set to 2. A linear warmup for the first 6\% of steps followed by a linear decay to 0 is used. To avoid the exploding gradient problem, the gradient norm is clipped to within 1. All the texts are tokenized using Byte-Pair Encoding(BPE) \cite{2015arXiv150807909S} and are chopped to sequences no longer than 512 tokens. Layer-wise decay rate $\xi$ is 0.9 and loss weight $\beta_1=5, \beta_2=\beta_3=\beta_4=1$.

\subsection{Ablation}
To assess the impact of each method we apply, we perform a series of analyses and ablations. The methods are added one by one to the baseline model. As shown in Table \ref{ablation}, Rationale Tagging multi-task and Adversarial Training bring relatively significant improvements. With all the methods in Table \ref{ablation}, the F1 score of our model (single) in dev-set attains to 91.3.

\begin{table}[t!]
\begin{center}
\begin{tabular}{|l|c|}
\hline \bf Model & \bf In-domain \\ \hline
Baseline Model & 89.5 \\
+ Rationale Tagging Task & 90.0  \\
+ Adversarial Training & 90.7 \\
+ Knowledge Distillation & 91.0 \\
+ Post-Processing & 91.3 \\
Ensemble & 91.8 \\
\hline
\end{tabular}
\end{center}
\caption{\label{ablation} Ablation study on the CoQA dev-set}
\end{table}

\subsection{Results}
We submit our system to the public CoQA leaderboard and compare its performance with others on the CoQA test set in Table \ref{result}.
Our system outperforms the current state-of-the-art single model by a 2.6\% absolute improvement in F1 score and scores 90.7 points after ensemble. In addition, while most of the top systems rely on additional supervised data, our system do not use any extra training data. 

\begin{table}[t!]
\begin{center}
\begin{tabular}{|l|c|}
\hline \bf Model & \bf F1 bound \\ \hline
our system & 91.8 \\
$\text{upper bound}_{1st}$ & 93.0 (+1.2) \\
$\text{upper bound}_4$ & 95.1 (+3.3) \\
\hline
\end{tabular}
\end{center}
\caption{\label{bound} Upper bound analysis on the CoQA dev-set}
\end{table}

\begin{table*}[t!]
\centering
\begin{tabular}{|p{3cm}|p{3.5cm}|p{2.5cm}|p{3cm}|p{2cm}|}
\hline
\bf Question & \bf Rationale & \bf Ground truth & \bf Our Answer & \bf Error Type \\
\hline
How many factors contribute to endemism?& Physical, climatic, and biological factors & Three  & Physical, climatic, and biological factors & Counting \\
\hline
Who told her? & Your mother told me that you had a part-time job & Sandy's mother & mother & Pronoun Paraphrasing\\
\hline
What did she do after waving thanks? & while driving off in the cab. & Drove off in the cab  & driving off in cab & Tense Paraphrasing \\
\hline
When? & Not till I spoke to him & When he spoke to him  & spoke to him & Conjunction Paraphrasing\\
\hline
\end{tabular}
\caption{\label{badcase} Some typical types of bad cases.}
\end{table*}

\section{Analysis}
In this section, a comprehensive analysis explains the limitation of extracitve models and our system. To figure out the headroom of extractive models for improvement, we first compute the proportion of free-form answers and then estimate the upper bound of extractive models in F1 score. At last, we analyze some typical bad cases of our system.

\subsection{CoQA}
 CoQA \cite{2018arXiv180807042R} is a typical conversational question answering dataset which contains 127k questions with answers obtained from 8k conversations and text passages from seven diverse domains. 
 
 An important feature of the CoQA dataset is that the answer to some of its questions could be free-form. Therefore, around 33.2\% of the answers do not have an exact match subsequence in the given passage. Among these answers, the answers Yes and No constitute 78.8\%. The next majority is 14.3\% of answers which are edits to text span to improve fluency. The rest includes 5.1\% for counting and 1.8\% for selecting a choice from the question.

\subsection{Upper Bound of Extractive Models}
Considering the answers could be free-form text, we estimate the upper bound of the performance of extractive method. For each question, the subsequence of the passage with the highest F1 score is regarded as the best answer possiblely for extractive models. As shown in Table \ref{bound}, with the first human answer as reference, the upper bound of F1 is 93.0. With all 4 human answers considered, the F1 can reach 95.1, indicating that the headroom for generative models is only 4.9\%. This is also the reason why we use an extractive model.

\subsection{Error Analysis}
As shown in Table \ref{badcase}, there are two typical types of bad cases in our system, e.g. Counting and Paraphrasing. To solve these problems, the model must have the capability to paraphrase the answer according to its question format. However, since cases of these two types only account for a small proportion of the entire data set, introducing a naive generative mechanism may have a negative effect, especially for the questions whose answers could be found in the passage. 

\section{Future directions}
While it is shown that our model can achieve state-of-the-art performance on CoQA dataset, there are still some possible directions to work on. One possible direction is to combine the extractive model and generative model in a more effective way to alleviate the free-form answer problem mentioned above. Another direction is to incorporate syntax and knowledge base information into our system. Furthermore, proper data augmentation may be useful to boost model performance.  


\bibliography{ref}
\bibliographystyle{acl_natbib}

\end{document}